\documentclass[conference,compsoc]{IEEEtran}

\usepackage{epsfig,endnotes}
\usepackage{balance}  
\usepackage{graphics} 
\usepackage{subfigure}
\usepackage{epstopdf}
\usepackage{float}
\usepackage{tabularx}
\usepackage{paralist}
\usepackage[usenames,dvipsnames]{xcolor}
\usepackage{amsmath}
\usepackage{algorithm}
\usepackage[noend]{algpseudocode}

\usepackage{times}
\usepackage{latexsym}
\usepackage{xspace}

\usepackage{csquotes}
\usepackage{multirow}
\usepackage{graphicx}
\usepackage{tikz}
\usepackage{tikz-3dplot}
\usepackage{amssymb}
\usepackage{stackengine,inlinenum}

\usepackage{url}
\makeatletter
\g@addto@macro{\UrlBreaks}{\UrlOrds}
\makeatother

\newcommand{\eg}{e.\,g.,\ }

\begin{document}
\title{Review Helpfulness Assessment based on Convolutional Neural Network}


\author{
\IEEEauthorblockN{Xianshan Qu, Xiaopeng Li, John R. Rose}
\IEEEauthorblockA{Dept. of Computer Science and Engineering\\
University of South Carolina, Columbia, SC 29205\\
Email: {xqu, xl4, rose}@cec.sc.edu}}

\maketitle
\begin{abstract}
In this paper we describe the implementation of a convolutional neural network (CNN) used to assess online review helpfulness. To our knowledge, this is the first use of this architecture to address this problem. We explore the impact of two related factors impacting CNN performance: different word embedding initializations and different input review lengths. We also propose an approach to combining rating star information with review text to further improve prediction accuracy. We demonstrate that this can improve the overall accuracy by 2\%. Finally, we evaluate the method on a benchmark dataset and show an improvement in accuracy relative to published results for traditional methods of 2.5\% for a model trained using only review text and  4.24\% for a model trained on a combination of rating star information and review text.
\end{abstract}

\section{INTRODUCTION}
\label{sec:introduction}

With the rapid growth of the internet, e-commerce is becoming an important part of life. According to a recent study~\cite{survey1}, more than half of Americans prefered to shop online in 2017. To improve customer experience, most shopping sites allows their customers to comment and rate purchased products. The reviews are helpful for potential customers to select products from the huge number of options available~\cite{Zhu_2010, Mudambi_2010, Baek_2012}. However, the quality of reviews varies greatly and the large number of available reviews can result in information overload for consumers.  In order to help users quickly find useful reviews, some platforms have implemented a mechanism that allows users to give feedback on the helpfulness of reviews. For example, on Amazon each review is usually accompanied with information indicating how many people found the review helpful, e.g., ``\textit{33 of 34 people found the following review helpful}''. Yet, such information is not available when it comes to the reviews recently posted. In addition, reviews with very few votes are often not trusted. Consequently, there is a need to be able to automatically evaluate the helpfulness of all reviews. 


Convolutional neural networks (CNN) have achieved major breakthrough in image classification and speech recognition~\cite{Krizhevsky_2012,Graves_2013}. They have also been applied to solve some natural language processing (NLP) problems~\cite{Goldberg_2015,Moreno_2017}, such as document classification~\cite{NIPS2015_zhang, AAAI159745, Johnson_2015, Yang2016HierarchicalAN, Hoa_2017}, sentiment analysis~\cite{Kim14f}, sentence modeling~\cite{KalchbrennerACL2014, Kim_2016} and have achieved promising results. The convolutional and pooling layers in a CNN architecture are able to capture local information from text and extract the most important features. This is our motivation for implementing and evaluating a CNN to assess the helpfulness of reviews. 

Previous studies~\cite{Kim_2006, Liu_2007, Hong_2012} have indicated that the helpfulness of a review can be affected by its content as it may contain product attributes and a description of the product that other customers may be interested in. Consider the following example:
\begin{displayquote}

The mouse itself is quite \emph{basic} \ldots \\
\ldots \,is \emph{affordable}, \emph{portable}, and very \emph{easy to use} \ldots

\end{displayquote}
This review describes many aspects of a mouse and would be considered helpful to customers who are concerned with those attributes.

%
%

In addition to the content of a review, the sentiment of a review can also affect the perceived helpfulness of a review~\cite{Mudambi_2010, MALIK2017290}. Customers may have different expectations for reviews that express different sentiments. For example, a negative review can be considered helpful with the description of one single bad aspect of the product. Consider the review for a mouse shown below:

\begin{displayquote}

The optical on this mouse died after four months of occasional use \ldots

\end{displayquote}
In contrast, customers may expect positive reviews to contain a thorough description of the product covering many different product aspects. Thus, the perceived helpfulness of a review is affected by both review content and sentiment.

Motivated by the preceding observations, we present a CNN-based architecture that incorporates both \textit{\textbf{review content}} and \textit{\textbf{review sentiment}} to assess review helpfulness. We use rating star information as a proxy for sentiment. The main contributions are summarized as follows:
\begin{itemize}

	\item We implement a CNN architecture to assess review helpfulness. To our knowledge, this is the first use of this architecture to address this problem.

	\item Inspired by the idea that rating star information and review text may affect the helpfulness of a review synergistically, we propose and evaluate two approaches to combining rating star information with review text for training a CNN. We report an increase in accuracy of 2\% compared to that of models trained only on review text.

	\item We evaluate the architecture on two datasets. We use a large dataset for exploring different factors affecting model performance. We also evaluate our approach on a small dataset for which published results exist. Our method can increase accuracy relative to previous results by 2.5\% for a model trained only on review text and 4.24\% for a model trained on a rating star information and review text.
\end{itemize}
\section{Related Work}
\label{sec:related_work}
In this section, we introduce related work from two aspects: review helpfulness assessment and applications of deep learning methods to natural language processing problems.

\subsection{Review Helpfulness Assessment}
Previous work in assessing review helpfulness has concentrated on using different machine learning algorithms and mining useful features from either the content or the ``metadata'' of the reviews (\eg the reviewers who wrote the reviews, the products that the reviews are written for, etc.). \cite{Kim_2006}  designed a system based on SVM regression to rank review helpfulness. They investigated a variety of features from Amazon product reviews, and found that features such as review length, unigrams and product ratings are most useful in measuring review helpfulness. \cite{Liu_2007} trained classification models based on SVMLight toolkit in order to detect low-quality reviews. They studied features from the sentence and word level of reviews and also the product level. Their models are tested on a self-annotated dataset consisting of 4909 reviews. \cite{Hong_2012} implemented a LIBSVM classifier to distinguish helpful from unhelpful reviews and an SVM-based helpfulness ranking system. In addition to using the textual features from reviews, they improved their system by extracting features from user preferences that could serve as the clues to the opinions of users on review helpfulness.


\begin{figure*}[h]
	\graphicspath{ {./figures/} }
	\centering
	\includegraphics[width=0.94\linewidth]{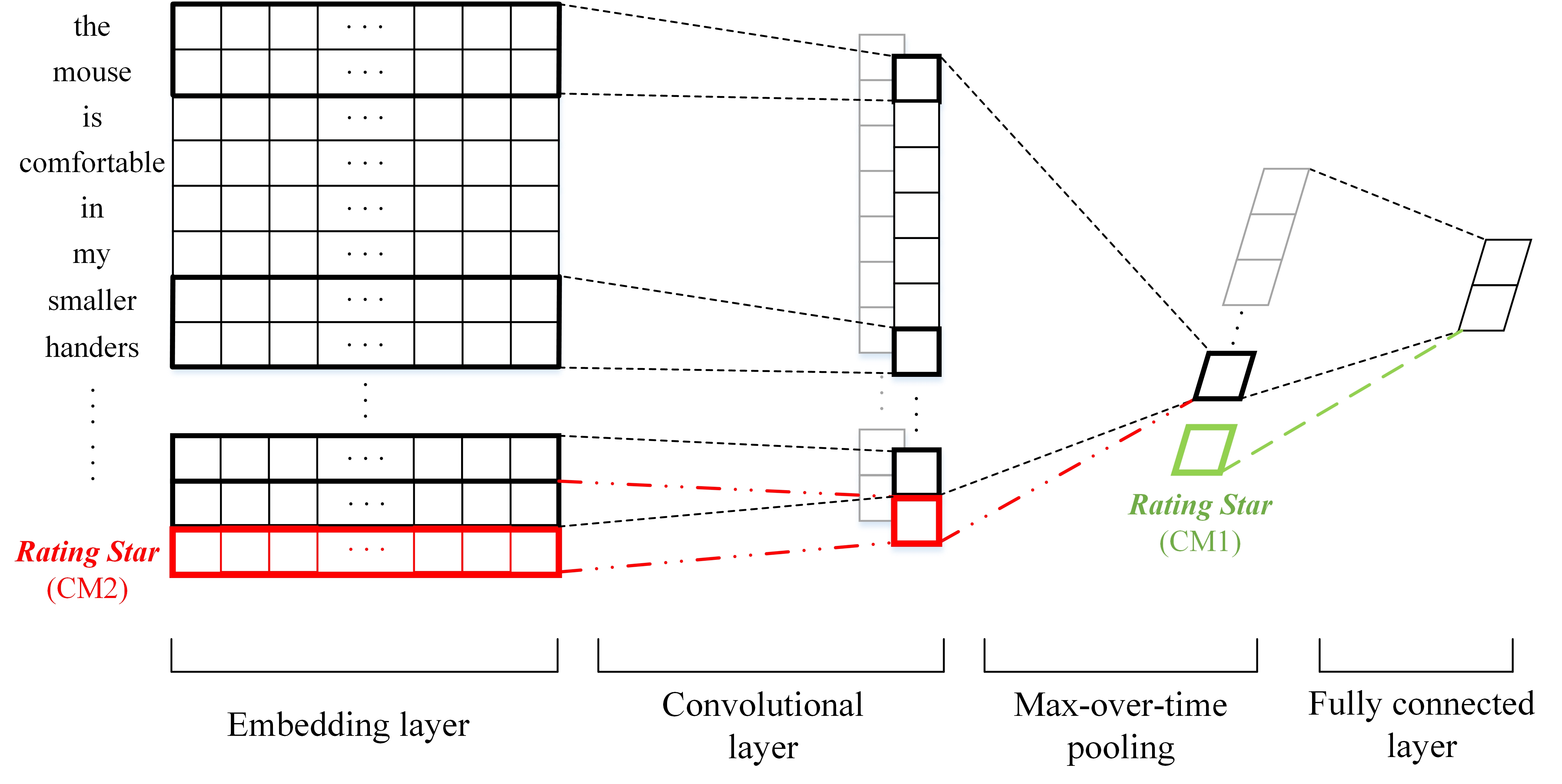}
	\caption{Architecture of CNN. Black color refers to the implemented CNN, green color denotes the first combination approach (CM1), and red color indicates the second combination approach (CM2)}
	\label{fig:archi_CNN}
\end{figure*}

\subsection{Deep Learning in NLP}
In contrast to previous studies, we assess review helpfulness using deep learning models. In the area of NLP, many researchers have shown that deep learning methods are effective on a variety of tasks. \cite{Kim14f} explored the applications of a simple CNN built on pre-trained word vectors to sentence classification. This paper reported that a simple CNN with one layer of convolution performs well on four out of seven datasets. \cite{NIPS2015_zhang} considered text as a kind of character-level signal and applied CNNs on characters for text classification. They achieved competitive results on six large-scale datasets. \cite{AAAI159745} introduced a recurrent convolutional neural network for text classification. Their model can capture the contextual information in the text by applying a recurrent structure. \cite{Yang2016HierarchicalAN} proposed a hierarchical network which builds the document vector from word vectors for document classification. They also introduced a mechanism to extract the words and sentences that contribute most to the classification.

\section{Methods}
\label{sec:model}

We first review the CNN architecture that we employ and then introduce factors that affect classification performance when assessing review helpfulness. Then we introduce two approaches to combining rating star information with review text in the CNN architecture.

\subsection{Convolutional Neural Network}
\cite{Hoa_2017} have demonstrated that a shallow-and-wide network at the word-level is currently the most effective method for some natural language problems, in particular for text classification and sentiment analysis. In this paper, we choose a classic shallow-and-wide CNN proposed by~\cite{Kim14f} to address the problem of review helpfulness assessment. 

As depicted in Figure~\ref{fig:archi_CNN}, the CNN consists of four layers: an embedding layer, a convolutional layer, a pooling layer and a fully connected layer. In the embedding layer, each review is embedded at the word-level and is represented as a matrix with each row corresponding to a word. In the convolutional layer, the width of a filter is usually fixed to be the same as the dimensionality of a word vector in order to capture the relation among adjacent words.
Sliding the filter along a row produces a feature map. Then a max-over-time pooling operation is applied to extract the maximum value corresponding to each feature map. The idea behind that is to capture the most important feature from each feature map. The pooling scheme can also be used to deal with the varying length of the feature maps produced by filters of different sizes. From preceding layers, features are extracted from filters and concatenated in the fully connected layer to produce a probability distribution over output classes. 

\subsection{Choice of Word Embedding Initialization}
When evaluating the CNN architecture for our task, there are two factors that impact model performance. The first factor to consider is the choice of word embedding initialization. In order to train with review text at the word-level, we must first convert each word into a vector representation. There are primarily two different initialization approaches of word embedding. One approach is to use pre-trained word vectors such as GloVe~\footnote{https://nlp.stanford.edu/projects/glove} (Global Vectors) which is a publicly available set of word vector initialization values from the Stanford NLP Group. GloVe have been pre-trained on word-word co-occurrence statistics from using several corpora. They produce a vector space with meaningful substructure. The other approach is to use randomized values. \cite{Kim14f} has suggested that for some text classifcation and sentiment analysis tasks, initialization with pre-trained vectors and learning specific tasks through fine tuning can achieve better results than using randomized initialization. In this paper, we explore this issue by training with two different word embedding methods. This is described in detail in Sections~\ref{sec:experiments}.

\subsection{Choice of Input Review Length}
Another factor that needs to be explored is the choice of length of input review text. For online reviews, there are large variations in review lengths. This is especially true of book reviews with a possible minimum length of 10 words and a maximum length of 5000 words in our datasets. In previous NLP tasks, the classification is either based on sentences where the maximum length is limited, or based on truncated reviews where the first few words are used. In the case of assessing review helpfulness, it is possible for important sentences to locate at the end of long reviews. We therefore evaluate the impact of truncated reviews of different fixed-length on model accuracy. Experiments addresses this issue are presented in Section~\ref{sec:experiments}.

\subsection{Combination Methods}
Additional types of information have been proven to be effective to assessing the helpfulness of reviews~\cite{Mudambi_2010}. Unfortunately, some adjunct information may not be available for all online reviews (e.g., information characterizing reviewer expertise). Also, some information may require explicit extraction (e.g., product-specific aspects from review texts). In contrast, rating star information is almost always available for each online review and does not require any specific extraction method. This makes this rating star information more applicable for large datasets. In this paper we propose two different approaches to incorporating rating star information.

The first approach is based on the ituition that rating star information is an explicit feature. This is how it would be handled in a traditional method for this problem. In this case it is simply added to the fully connected layer of the CNN. This approach is identified as CM1 in figure 1. In the fully connected layer, 
each output node is a weighted composition of the input, and a softmax function will be applied on each output node to get a probability between 0 and 1 as shown in Eq.~\ref{eq:softmax},
\begin{equation}
\mathrm{P}(y=j|\mathbf{x}) = \frac{e^{\mathbf{x}^{T}\mathbf{w}_{j}}} {\sum_{k=1}^{K} e^{\mathbf{x}^{T}\mathbf{w}_{j}}} \,,
\label{eq:softmax}
\end{equation}
where $x$ is the input vector that contains the rating star information, and $P$ is the probability for each class. This equation defines the effect of rating star information on the ouput. The degree of impact is decided by the weight $w$ assigned to it.

The second approach to incorporating rating star information is inspired by the observation that humans evaluate the review text and the rating stars at the same time. In this apporach we combine the rating star information with review text at the very begining to capture the relation thoroughly. This approach is identified as CM2 in figure 1. 

We convert the rating star information into a vector with the same dimensionality of the word in the embedding layer. We then concatenate it at the end of the representative matrix of a review. For example, a review of length n (padded if necessary) with a rating star at the end can be represented as
\begin{equation}
\mathbf{x}_{1:n+1} = \mathbf{x}_1 \oplus \mathbf{x}_2  \cdots \oplus \mathbf{x}_n \oplus \mathbf{x}_{n+1} \,,
\end{equation}
where $\mathbf{x}_{i} \in \mathbb{R}^k$, $\mathbf{x}_{1}, \mathbf{x}_{2}, \cdots ,\mathbf{x}_{n}$ refer to review words, $\mathbf{x}_{n+1}$ indicates the vector embedding of a rating star and $\oplus$ refers to the concatenation operator. As there are 5 rating stars (1 to 5) for a review, we consider them as 5 new words added to the vocabulary learned from the review corpus. In the convolution operation, a filter $\mathbf{w} \in \mathbb{R}^{hk}$ is applied on a window of $h$ words and produces a feature $c_i$ as shown in Eq. \ref{eq: feature_produce}, 
\begin{equation}
c_i = f(\mathbf{w} \cdot \mathbf{x}_{i:i+h-1} + b) \,,
\label{eq: feature_produce}
\end{equation}
where $\mathbf{x}_{i:i+h-1}$ refers to the word window, $b$ is a bias term, and $f$ is a non-linear activation function.
The filter is applied to each possible word window and produces a feature map as shown in Eq. \ref{eq: convolution}. 
\begin{equation}
\mathbf{c} = (c_1, c_2, \cdots , c_{n-h+2}) \,,
\label{eq: convolution}
\end{equation}
where the last item $c_{n-h+2}$ contains information of the rating star.
During the training process, the filter can learn the relation among neighboring words and rating stars. 
The most important features will be extracted through the max-over-time pooling operation $\hat{c}=\max \{ c \}$, and both the features from review text and rating stars will be taken into consideration. 

\section{Experiments}
\label{sec:experiments}
We conducted experiments on two datasets. We used a large dataset for training models and exploring different factors to determine their impact on model acccuracy. We also used a small dataset for which published results are available in order to compare the performance of our models with that of traditional methods. 
\subsection{Datasets}

\begin{table}[t]
  \centering
  \begin{tabular}{cc}
    \hline
    \textbf{Category}  &  \textbf{Review}  \\ \hline
	Electronics  &  288,706  \\ \hline 
    Books  &  895,074  \\ \hline  
  \end{tabular} 
  \caption{Number of reviews of each product type after the preprocessing.}  
  \label{tab:review_number} 
\end{table} 

\textbf{Large dataset.} We obtained an Amazon review dataset from Amazon product data~\cite{He_2016}.~\footnote{http://jmcauley.ucsd.edu/data/amazon}. This dataset includes 142.8 million reviews covering 24 product types and spans a period of 18 years. The reviews have product names, rating stars, review texts, helpfulness votes and total votes. From the different product types which range from books to office products, we chose two representative product types: Books and Electronics. These two categories have the most reviews and have been frequently studied before.

\textbf{Small dataset.} This dataset contains 34266 Amazon reviews most of which were released in 2006, and cover 16 different categories including electronics and books. This dataset was provided by Yu Hong~\footnote{http://nlp.suda.edu.cn/~hong/}. He and his colleagues have implemented several traditional methods for assessing review helpfulness  and have reported the testing accuracy for these methods~\cite{Hong_2012}. We will use this dataset as the benchmark dataset for comparison with state-of-the-art results.

\textbf{Data preprocessing.} The small dataset has been preprocessed and labelled. For the large dataset, we first filtered out reviews with fewer than 6 total votes since such reviews may have a biased helpfulness voting ratio (i.e., the ratio of helpful votes to total votes) ~\cite{Kim_2006}. We then removed  duplicated reviews by matching bigrams between each pair of reviews. A pair of reviews are deemed duplicates if more than 80\% of the bigrams match~\cite{Kim_2006, Hong_2012}. Since we trained our models on the large dataset and evaluated them on the small dataset, we also removed reviews appearing in both datasets in order to avoid overlaps that might skew testing result. 

\begin{figure}[t]
	\graphicspath{ {./figures/} }
	\centering
	\includegraphics[width=0.92\linewidth]{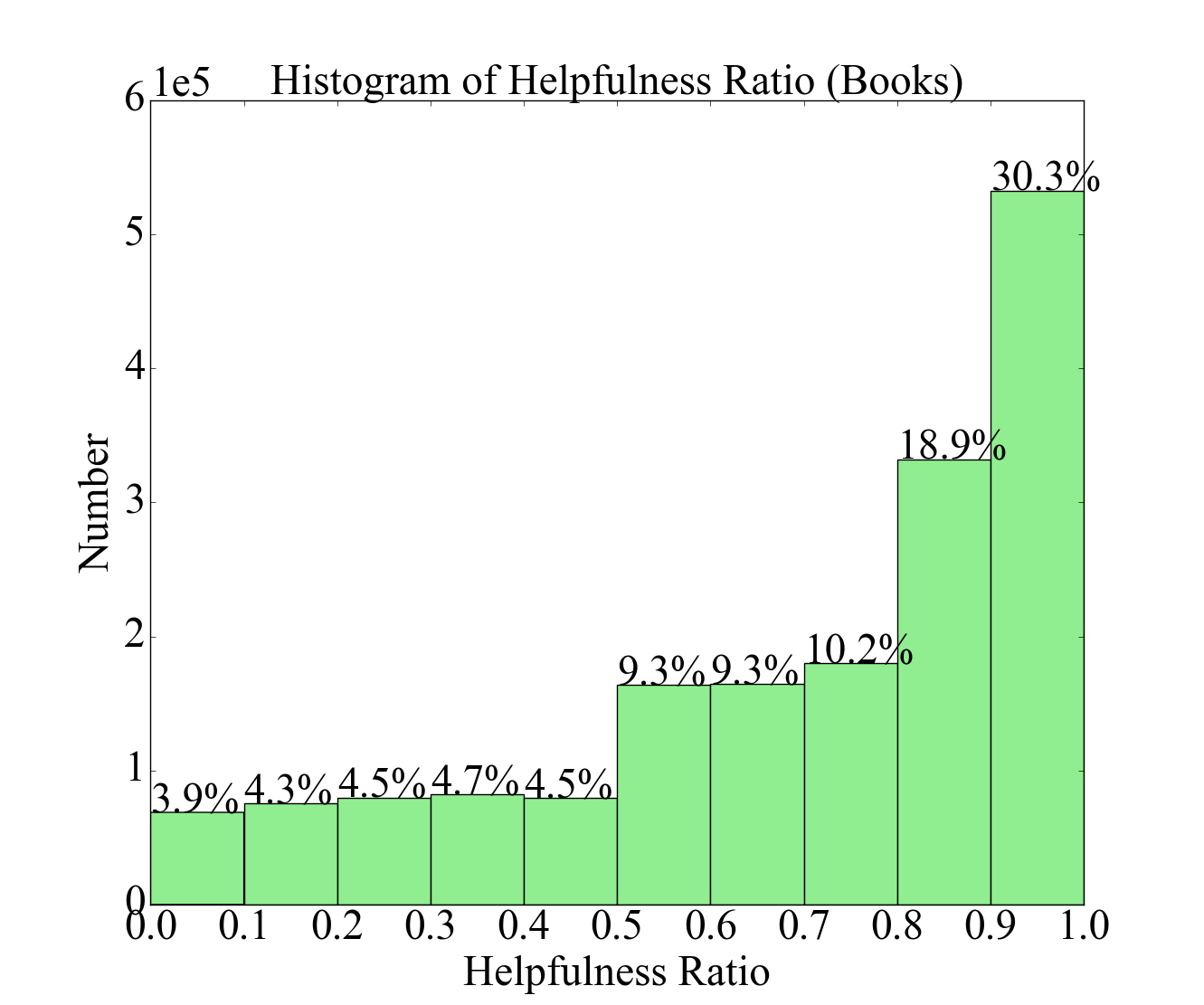}
	\caption{The histogram of helpfulness voting ratio of book reviews.}
	\label{fig:hist_help_ratio}
\end{figure}

%

In order to label each review as helpful or unhelpful, it is necesary to first  select a voting ratio threshold. We plotted a histogram of the helpfulness voting ratio of the large dataset. As shown in Figure~\ref{fig:hist_help_ratio} (the histogram for electronic reviews is similar and not included due to space limitation), the helpfulness ratio is equally distributed within the range from 0 to 0.5, while after 0.5 there is a sudden jump. Because of the natural separation of review helpfulness ratio at 0.5, we choose 0.5 as the boundary. Reviews with a helpful voting ratio above 0.5 are labelled as helpful, the remaining reviews are labelled as unhelpful.
It turns out that using a helpfulness ratio of 0.5 on this dataset results in many more reviews being labelled as helpful than unhelpful.
We randomly filtered out some helpful samples to make sure that the number of samples in two classes (helpful and unhelpful) is balanced. The final large dataset is summarized by category in Table \ref{tab:review_number}.

\subsection{Experiments with large dataset}
When employing the convolutional neural network described in Section \ref{sec:model} on the large dataset, we first explored models trained with review text only. We then evaluated models trained with a combination of review rating stars with review text. We considered two different approaches to combining  review rating stars and review text. This is described as follows.

\textbf{Training only with review text.} Our first experiments focused on train a convolutional neural network using only review text. In order to explore and identify factors which may affect performance, we use the same identical dataset for the series of tests in this section. We divided the dataset into 10 equal-sized partitions. Of these, eight of the partitions were used for training, one partition for validation, and the last partition for testing. 

We explored the impact of the following two factors:
\begin{itemize}
\item Different initializations of word vectors
\item Different input lengths of the review words
\end{itemize}
We first investigated the effect of using different methods to initialize the word vectors. Aside from initialization, all other aspects of the imlementation were kept the same. The first initialization approach is to use GloVe which have dimensionality of 100. 
The other approach to initialization that we investigated was to use randomized values between -0.05 and 0.05 to initialize each dimension of the word vectors and keep the same word embedding dimensionality of 100. In both cases, the word vectors will be optimized during the training process.
\begin{figure}[t]
	\graphicspath{ {./figures/} }
	\centering
	\includegraphics[width=0.92\linewidth]{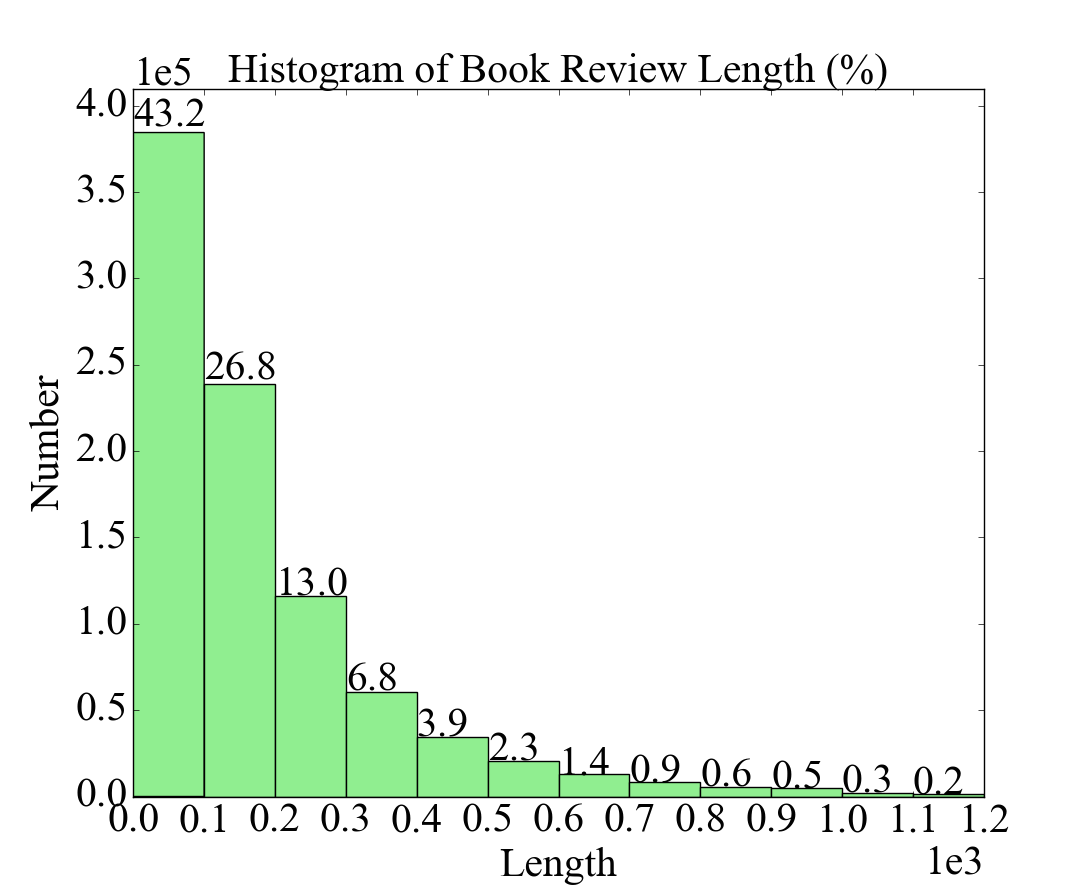}
	\caption{The histogram of book review length (number of words).}
	\label{fig:hist_length}
\end{figure}

%

We next explored the effect of different input review lengths. We started by plotting histograms to visualize the distribution of review lengths. The length of a review is expressed in the number of words in a review. Based on Figure \ref{fig:hist_length} (the histogram for electronic reviews is similar and not included due to space limitation), we find that the first 3 bins (300 words) account for around 80\% of review lengths. We therefore investigated fixed lengths of 50, 100, 200, and 300 words. In experiments we iterated through each of the fixed input lengths in order to identify the effect that input length has on performance. We also included the length of 1200 words for comparison as this length includes 99\% of reviews in the dataset. When the real review length is shorter than the fixed input length, we pad the review with zeros. Conversely, when the real review length is longer than the fixed length boundary, we truncate the review. Keeping all other aspects same, we trained models with different input review lengths in order to identify the impact of review length on model accuracy.

\textbf{Training with review text and rating star.} We next considered how including rating star information might improve classification performance. As described in Section \ref{sec:model}, we investigated two approaches. In order to compare the results of these two combination methods, we performed a 10-fold cross validation. At each iteration, eight of the partitions form the training set, one partition acts as a validation set and one partition acts as a test set. Each partition participates exactly once as the validation set and once as the test set.

In our experiments, we use accuracy and F1-score to describe the performance of the helpfulness-based review classification. Accuracy is the ratio of the reviews whose helpfulness is correctly determined. The F1-score is the harmonic mean of the precision and recall. 

\subsection{Experiments with small dataset}
Due to the limited amount of training data in the small dataset, we started with models initially trained using large dataset. We then fine tuned the parameters relative to the small dataset by performing a 5-fold cross validation with the small dataset. The two starting models (one based on reviews, and the other based on reviews and rating stars) were trained using book reviews.

\section{Results and Discussion}
\label{sec:Results}

\subsection{Results for the large dataset}
\begin{figure}[h]
	\graphicspath{ {./figures/} }
	\centering
	\includegraphics[width=0.94\linewidth]{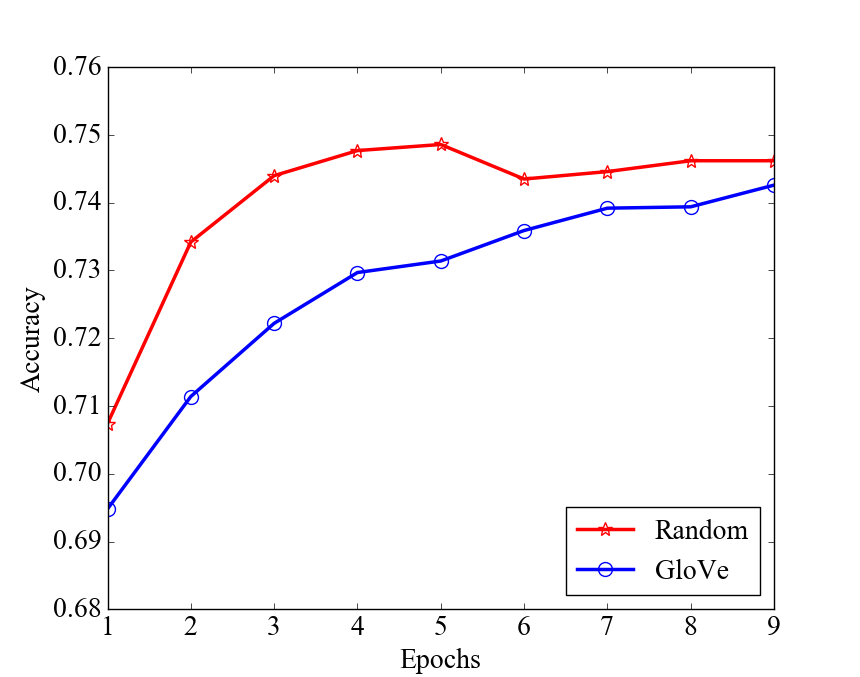}
	\caption{The effect of two word embedding initializations (Random vs Glove) on model accuracy.}
	\label{fig:random_glove}
\end{figure}

\textbf{Effect of different word embedding initializations.} The model accuracies resulting from the two different approaches to initializating word vectors is shown in Figure \ref{fig:random_glove}. Our initial expectation had been that initilization using pre-trained word vectors would start from a good point, and thus work better than randomized initlization. However, as is evident in Figure \ref{fig:random_glove}, the results indicate a very slight difference in final accuracy for two different initialization methods.  This result is different from previously published results~\cite{Kim14f}. There may be various reasons for the outcome. First, our problem domain differs from theirs, thus the trained model may not perform well for our problem. More importantly, we have many more training samples. Consequently,  word vectors initialized with small randomized values can be trained well enough and may converge more quickly. This can be inferred from the two validation curves during training process as shown in Figure \ref{fig:random_glove}. After the first epoch, the validation accuracy of the randomized initialization is higher than that of GloVe initialization, and it reaches the highest accuracy after training of 5 epochs. In contrast, the model initialized with GloVe vectors reach the same level of validation accuracy after around 9 epochs of training. These results suggest that a speed up of the training process can be achieved by randomly initializing the word vectors when enough training data is available.

\begin{figure}[h]
	\graphicspath{ {./figures/} }
	\centering
	\includegraphics[width=0.94\linewidth]{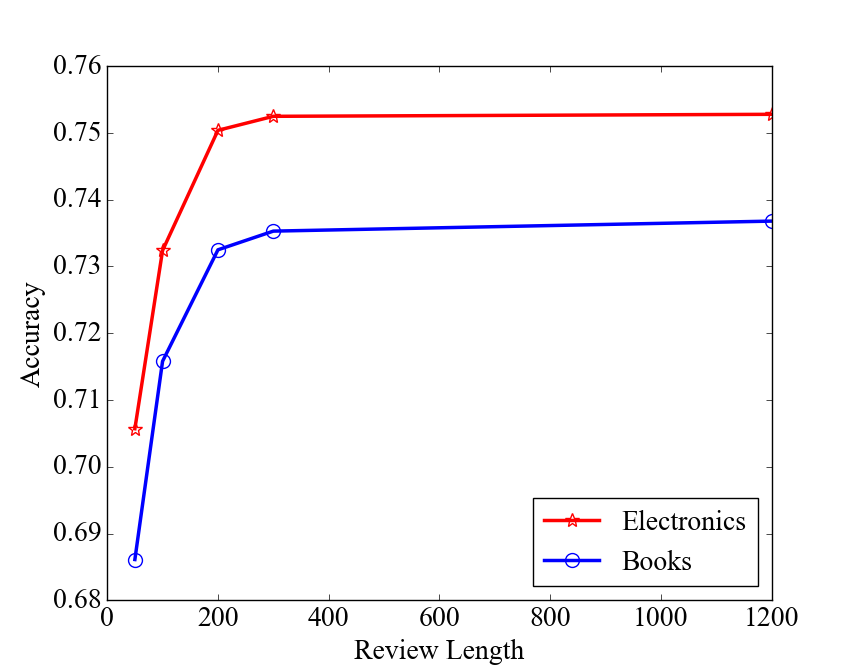}
	\caption{The effect of review length on model accuracy.}
	\label{fig:length_accuracy}
\end{figure}

\begin{table*}[ht]
  \centering
  \begin{tabular}{*7c}
    \hline
    \multirow{2}{*}{\textbf{Category}}  &  \multicolumn{2}{c}{\textbf{Review Text}}  &  \multicolumn{2}{c}{\textbf{CM1}} & \multicolumn{2}{c}{\textbf{CM2}}  \\
    \cline{2-7}
       &  Accuracy(\%)  &  F1(\%)  &  Accuracy(\%)  &  F1(\%)  &  Accuracy(\%)  &  F1(\%)  \\ \hline
	Electronics  &  74.82  &  75  &  75.05  &  75  &  76.96  &  77  \\
    Books  &  73.22  &  73  &  73.45  &  73  &  75.17  &  75  \\ \hline  
  \end{tabular}
  \caption{Testing results on the large dataset including elctronics and books.}
  \label{tab:our_model_result}
\end{table*}

\begin{table*}[ht]
  \centering
  \begin{tabular}{*3c}
    \hline
    \textbf{Model}  &  \textbf{Accuracy}(\%)  &  \textbf{F1}(\%)  \\ \hline
    \cite{Kim_2006}  &  61.29  &  -  \\
    \cite{Liu_2007}  &  62.85  &  -  \\
    \cite{Hong_2012}  &  69.62  &  -  \\
    Combined Features~\cite{Hong_2012}  &  71.9  &  -  \\
    CNN\_{}Text  &  74.4  &  74.4  \\
    CNN\_{}Text\_{}Star  &  76.14  &  76  \\ \hline  
  \end{tabular}
  \caption{Comparision with previous studies on the small dataset.}
  \label{tab:comparison_small_dataset}
\end{table*} 

\textbf{Effect of different input review lengths.} Figure \ref{fig:length_accuracy} shows the effect of review length on model accuracy. The models were trained using different fixed input review lengths ranging from 50, 100, 200, 300 to 1200 words. As shown in this figure, increasing the review length from 50 words to 100 words results in an increase of accuracy of over 2\%. Doubling the review length from 100 words to 200 words also results in a significant increase in accuracy. Relative to models derived from reviews of 50 words, models derived from reviews of 200 words display more than 4\% increase in accuracy. However, models derived from review lengths of more than 200 words show diminishing improvement. While there is a slight increase of 0.2\% from 200 to 300 words, the difference in accuracy between models trained on review lengths of 300 words and 1200 words is negligible.
This result suggests that it may not be advisable to use the maximum review length as the input review length. By restricting the reviews to the first 300 words for training, we can not only maintain the same level of accuracy as that achieved with the maximum review length, but also make the training process 4 times quicker.


\textbf{Effect of including rating star information.} As described in Section \ref{sec:model}, we evaluated two approaches to including rating star information in model development.  As shown in Table \ref{tab:our_model_result}, both approaches improve model accuracy to a varying degree. This suggests that rating star information is an important factor for assessing review helpfulness. In Table \ref{tab:our_model_result}, ``Review Text'' refers to models trained using only review text as input. The column labelled CM1 refers to the approach of using rating star information only after the network has extracted other features from the review text. The approach was only able to achieve an increase in accuracy of 0.23\%. However, the second approach (CM2) to incorporating rating star information achieves an improvement in accuracy of 2.14\% when compared to review text only based models. Recall that this approach includes rating star information as starting input to the convolutional neural network (CNN) and lets the CNN learn the proper feature representations.

From these results, we can conclude that combining rating star infomration with review text as initial input to the CNN produces the best results. It may because that the rating star information will have less of an effect on the prediction results if we put it before the fully connected layer. For example, if we use 16 filters for each filter size, and have 3 different filter sizes, then 48 features will be extracted from the review, and they will be used together with the rating star for prediction. We have observed that if we increase the number of filters, the increase of accuracy using CM1 will decrease correspondingly (figures not included due to space limitation). We hypothesize that this is due to the reduction in effect of rating star information relative to the increased number of extracted features from review text. In contrast, if the rating star information is combined with review text as input to the neural network from the begining, then the network will explore and extract important features without diluting the influence of the rating star information.

\textbf{Effect of different product types.} Table \ref{tab:our_model_result} presents a comparison of model accuracy for electronics and books. The results presented in this table do not suggest any obvious differences. The accuracy for books is a slightly lower than that of electronics, even though there are more training samples for books. The observed difference may be due to the observation that the book reviews tend to be more subjective, and thus have larger variance and need more training samples to achieve a high accuracy.

\subsection{Results for the small dataset}
We took the best model trained using the large dataset as a starting point and then fine tuned the parameters via a 5-fold cross validation on the small dataset. The results are shown in Table \ref{tab:comparison_small_dataset}. When training only with review text, we can obtain the accuracy of 74.4\%. When the network is trained with the combination of review text and rating star information, we achieve an accuracy of 76.14\%. Both of our results surpass the previously published best accuracy of 71.9\% achieved by combining all possible features~\cite{Hong_2012}. The convolutional neural network produces state-of-the-art results for this problem.

\section{Conclusion}
\label{sec:conclusion}
In this paper, we explore the possibility of employing a convolutional neural network to assess review helpfulness. During the process, we explore and analyze the impact of two related factors: word embedding initialization and input review length. From experiments on a large dataset, we conclude that randomized word embedding initlization performs as well as pre-trained GloVe initialization. We also conclude that training using the first 300 review words produces the same level of accuracy as that achieved with the maximum review length. 

Finally, we exlore the use of rating star information to further improve model acuracy. We propose and evaluate two approaches to combing rating star information with review text. We demostrate an increase in accuracy of 2\%. We then compare the performance of our method to that of previously published results. We start with a model trained on the large dataset used to evaluate training parameters. We fine tune that model be performing an subsequent 5-fold cross validation on the small dataset for which published results exist. We demostrate an increase in accuracy relative to previously published results of 2.5\% for a model trained using only review text and 4.24\% for a model trained on review text and rating star information.


\bibliography{sections/references}
\bibliographystyle{IEEEtran}


\end{document}